\documentclass[10pt,twocolumn,letterpaper]{article}

\usepackage{cvpr}
\usepackage{times}
\usepackage{epsfig}
\usepackage{graphicx}
\usepackage{amsmath}
\usepackage{amssymb}
\usepackage{enumitem}
\usepackage{booktabs}
\usepackage{xcolor}
\usepackage{multirow}
\usepackage{subcaption}
\usepackage{longtable}
\usepackage{adjustbox}

\usepackage[pagebackref=true,breaklinks=true,letterpaper=true,colorlinks,bookmarks=false]{hyperref}

\graphicspath{ {figures/} }

\usepackage[breaklinks=true,bookmarks=false]{hyperref}

\cvprfinalcopy 

\def\cvprPaperID{20} 

\ifcvprfinal\fi
\begin{document}

\title{Semi-supervised Learning: Fusion of Self-supervised, Supervised Learning, and Multimodal Cues for Tactical Driver Behavior Detection}

\author{Athma Narayanan\\
Honda Research Institute, USA\\
{\tt\small anarayanan@honda-ri.com}
\and
Yi-Ting Chen\\
Honda Research Institute, USA\\
{\tt\small ychen@honda-ri.com}\\
\and
Srikanth Malla\\
Honda Research Institute, USA\\
{\tt\small smalla@honda-ri.com}\\
}

\maketitle

\begin{abstract}
   In this paper, we presented a preliminary study for tactical driver behavior detection from untrimmed naturalistic driving recordings. While supervised learning based detection is a common approach, it suffers when labeled data is scarce. Manual annotation is both time-consuming and expensive. To emphasize this problem, we experimented on a 104-hour real-world naturalistic driving dataset with a set of predefined driving behaviors annotated. There are three challenges in the dataset. First, predefined driving behaviors are sparse in a naturalistic driving setting. Second, the distribution of driving behaviors is long-tail. Third, a huge intra-class variation is observed. To address these issues, recent self-supervised and supervised learning and fusion of multimodal cues are leveraged into our architecture design. Preliminary experiments and discussions are reported.          
\end{abstract}

\section{Introduction}
Intelligent transportation systems require interdisciplinary efforts including computer vision, machine learning, robotics, psychology, and control theory. It is challenging to drive in real world because decisions need to be made with incomplete information and diverse situations. Moreover, modeling uncertain behaviors of road users is still unsolved.

Towards this goal, we collected a naturalistic driving dataset, which will appear in CVPR'18 main conference~\cite{RamanishkaCVPR2018}
The total size of the dataset is 104 video hours with the predefined driving behaviors annotated. 
We defined a 4-layer scheme to annotate driver behaviors including tactical driver behaviors and interactive behaviors between the drivers and traffic participants. More details of the annotation and definition of driver behaviors will be provided in the supplementary material. Note that driving behaviors are a combination of driver behaviors, the interactive behaviors between driver and traffic participants, and traffic participants' behaviors. In this paper, we focus on detecting tactical driver behaviors as in~\cite{RamanishkaCVPR2018}. 

Manual annotation of driving behaviors is time-consuming and expensive. To minimize human efforts, automatic detection mechanism is necessary. While supervised learning is a common approach to address the problem, it suffers from when labeled data is scarce. This issue is presented in the collected dataset. Specifically, the dataset has the following three challenges. First, predefined driving behaviors are sparse. Only 15\% of data is labeled. Most of the time, drivers are doing "going straight," "stopping for red light," and "parking." Second, the distribution of driving behaviors is long-tail. For example, we observe more "turning" than "U-turn." Third, a huge intra-class variation is observed. For instance, a "turning right" is different from a "turning right while yielding a group of pedestrians."

We leverage recent advances in self-supervised learning for structure from motion~\cite{ZhouCVPR2017}, supervised learning for semantic segmentation~\cite{LinCVPR2017,ChenPAMI2018}, imbalance class distribution handling~\cite{Lin2017b}, and multimodal fusion~\cite{hazirbas2016fusenet} to address aforementioned issues. The proposed algorithm is presented.    




\setlength{\belowcaptionskip}{-10.5pt}

\begin{figure*}
\label{baseline2}
\begin{center}
\includegraphics[scale=0.21, trim={0cm 0cm 0cm 0cm},clip]{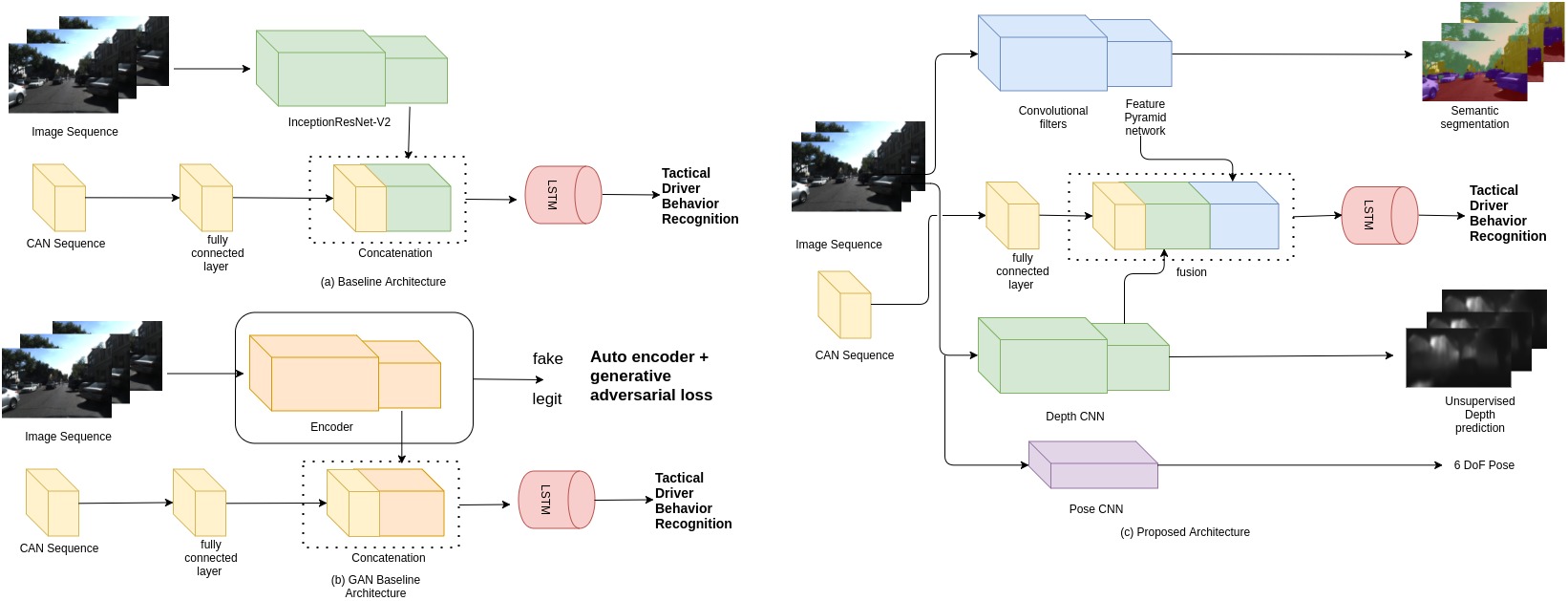}
   \caption{Figure (a) The baseline architecture combining InceptionResNet-V2 Image features and CAN sensor data~\cite{RamanishkaCVPR2018}. 
Figure (b) A baseline architecture using features obtained by~\cite{santana2016learning}. Figure (c) The proposed architecture combining self-supervised learning, semi-supervised learning and multimodal cues is presented for tactical driver behavior recognition.}
\end{center}   
\end{figure*}

\section{Methodology and Experiments}
 

We hypothesize that semantic context, 3D scene structure and vehicle motion are crucial tactical driver behavior detection. We intended to leverage features extracted from these cues than using features trained by supervision. This section gives the details of the architecture design and the preliminary results. 

Given a synchronized images and Controller
 Area Network (CAN bus) sensors data, the baseline model~\cite{RamanishkaCVPR2018} sampled input frames from video streams and values from CAN bus sensors at 3 Hz. The frame representation is extracted from the \textit{Conv2d\_7b\_1x1} layer of InceptionResnet-V2~\cite{SzegedyIV16} pretrained on ImageNet~\cite{Deng09imagenet}. The features are convolved with a $1 \times 1$ convolution to reduce
dimensionality from $8 \times 8 \times 1536$ to
$8 \times 8 \times 20$. Raw sensor values are passed through a fully-connected layer to obtain a one dimensional feature vector which is further concatenated with image features.
The concatenated features are fed into a Long-Short Term Memory (LSTM)~\cite{Hochreiter:1997:LSM:1246443.1246450} to encode the necessary history of past measurements. 
During training, we formed batches of sequence segments by sequentially iterating over driving sessions. The last LSTM hidden state from the previous batch is used to initialize the LSTM hidden state on the next step. The training is performed using truncated backpropagation through time. Addtionally, the data imbalance between foreground and background frames are handled using the recently proposed technique for modifying cross-entropy loss to deal with the class imbalance~\cite{Lin2017b}. Note that the details of the training protocol will be provided in the supplementary material.

%
%

Five different experiments were conducted as shown in Table \ref{table:results}. Note that we adopted the same architecture as in~\cite{RamanishkaCVPR2018} to detect tactical driver behaviors, but with different image features. First, we presented the baseline as in \cite{RamanishkaCVPR2018}. Second, we trained an auto-encoder with adversarial loss to generate image features as in \cite{santana2016learning}. We expect the reconstruction of images can learn the scene composition of the dataset. To reduce the reliance on direct supervision, we leveraged image reconstruction features to serve as a substitute. We took the intermediate encoder feature representation to the LSTM. Third, as 3D scene structure is crucial, we leveraged the unsupervised learning based structure from motion~\cite{ZhouCVPR2017}. 
%
%
Fourth, for semantic context, we modify Deeplab~\cite{ChenPAMI2018} to incorporate Feature Pyramid Network~\cite{LinCVPR2017} to enrich features at higher resolution features. Finally, features from 3D scene structure and semantic context are fused with CAN bus features by concatenation and batch normalization, similar to the work done in~\cite{hazirbas2016fusenet}. The aforementioned architectures are shown in Figure 1.

\begin{table}
\small
\caption{Experimental results on a set of 104-hour data. All number are in \%.} 
\centering 
  \begin{adjustbox}{max width=8cm}
\begin{tabular}{c c c c c c} 
\hline\hline 
Driver behavior class  &\cite{RamanishkaCVPR2018} & \cite{santana2016learning} & Depth+CAN & Seg+CAN & Our  \\ [0.5ex] 
\hline 
left lane change & 35.72 & 14.06& \textbf{38.96} &37.13&34.45  \\ 
right lane change & 25.48 & 6.42&25.32 &23.25&\bf{28.06} \\
railroad passing & \textbf{7.27} & 0.14& 0.82 &3.06&5.40  \\
left lane branch & 20.00 & 4.09 & 26.68 &35.94&\bf{43.05} \\
right lane branch & 0.74 & 0.59 &{1.58} &\bf{2.97}& {2.10} \\
left turn & 73.52 & 66.00&{74.21} &\bf{77.78}& 75.07 \\
right turn & 73.95 & 73.52 &\bf{76.20} &75.63&75.82 \\
U-turn & 15.78 & 26.77 &\bf{32.54} &\bf{27.77}&{26.40} \\
intersection passing & 74.12 & 29.45 &69.98 &\bf{79.69} &{77.70} \\
crosswalk passing & 4.04 & 0.63 &6.65 &\bf{14.02}&{13.14} \\
merge  & 6.35 & 0.40 &9.17 &14.83&\bf{16.42} \\[1ex] 
\hline
\textbf{mean} & 30.63 & 20.19 &32.92 &35.64&\textbf{36.15} \\[0.5ex]

\hline 
\end{tabular}
\end{adjustbox}
\label{table:results} 
\end{table}

\section{Discussion}
Our experiments indicate that robust fusion of image features from auxiliary tasks such as 3D scene structure and semantic context help the driver behavior detection tasks as demonstrated in Table \ref{table:results}. With  semantic context and 3D scene structure, we see improvements in classes such as intersection passing, cross-walk passing, U-turn and merge class. The proposed architecture improves the performance of~\cite{santana2016learning} by 16 \%. This demonstrates the effectiveness of the proposed features over features obtained by reconstruction. 

However, we expected a significant performance boost in those behaviors with strong correlations to semantic context (e.g., lane change due to the existence of lane markers in semantic context). The fusion of semantic context with CAN (i.e., $4^{th}$ column results) does not reflect this hypothesis. Note that \textbf{railroad} is not trained in the current semantic context algorithm. A better architecture design in the multimodal fusion, imbalanced distribution, and temporal modeling is necessary for further improvement.

\clearpage

{\small
\bibliographystyle{ieee}
\bibliography{egbib}

\begin{thebibliography}{10}\itemsep=-1pt

\bibitem{RamanishkaCVPR2018}
Anonymous.
\newblock Anonymous.
\newblock In {\em CVPR}, 2018.

\bibitem{ChenPAMI2018}
L.-C. Chen, G.~Papandreou, I.~Kokkinos, K.~Murphy, and A.~Yuille.
\newblock {DeepLab: Semantic Image Segmentation with Deep Convolutional Nets,
  Atrous Convolution, and Fully Connected CRFs}.
\newblock {\em PAMI)}, 40(4):834 -- 848, 2018.

\bibitem{Deng09imagenet}
J.~Deng, W.~Dong, R.~Socher, L.~jia Li, K.~Li, and L.~Fei-fei.
\newblock {ImageNet: A large-scale Hierarchical Image Database}.
\newblock In {\em CVPR}, 2009.

\bibitem{hazirbas2016fusenet}
C.~Hazirbas, L.~Ma, C.~Domokos, and D.~Cremers.
\newblock {Fusenet: Incorporating Depth Into Semantic Segmentation via
  Fusion-based CNN Architecture}.
\newblock In {\em ACCV}, 2016.

\bibitem{Hochreiter:1997:LSM:1246443.1246450}
S.~Hochreiter and J.~Schmidhuber.
\newblock {Long Short-Term Memory}.
\newblock {\em Neural Computation}, 9(8):1735--1780, 1997.

\bibitem{LinCVPR2017}
T.-Y. Lin, P.~Dollár, R.~Girshick, K.~He, B.~Hariharan, and S.~Belongie.
\newblock { Feature Pyramid Networks for Object Detection}.
\newblock In {\em CVPR}, 2017.

\bibitem{Lin2017b}
T.-Y. Lin, P.~Goyal, R.~Girshick, K.~He, and P.~Dollar.
\newblock {Focal Loss for Dense Object Detection}.
\newblock In {\em ICCV}, 2017.

\bibitem{santana2016learning}
E.~Santana and G.~Hotz.
\newblock {Learning a Driving Simulator}.
\newblock {\em arXiv preprint arXiv:1608.01230}, 2016.

\bibitem{SzegedyIV16}
C.~Szegedy, S.~Ioffe, and V.~Vanhoucke.
\newblock {Inception-v4, Inception-ResNet and the Impact of Residual
  Connections on Learning}.
\newblock {\em CoRR}, abs/1602.07261, 2016.

\bibitem{ZhouCVPR2017}
T.~Zhou, M.~Brown, N.~Snavely, and D.~G. Lowe.
\newblock {Unsupervised Learning of Depth and Ego-Motion from Video}.
\newblock In {\em CVPR}, 2017.

\end{thebibliography}
}

\end{document}